\documentclass[journal,twoside,web]{ieeecolor}
\usepackage{tmi}
\usepackage{cite}
\usepackage{amsmath,amssymb,amsfonts}
\usepackage{algorithmic}
\usepackage{graphicx}
\usepackage{textcomp}
\usepackage{bbm}
\usepackage{graphicx}

\begin{document}
\title{Data Adaptive Few-shot Multi Label Segmentation with Foundation Model}
\author{Gurunath Reddy M, Dattesh Shanbhag, and Deepa Anand
\thanks{Gurunath Reddy M, Dattesh Shanbhag, and Deepa Anand are with the Advanced Technology Group, GE HealthCare,
	Bangalore, India (e-mail: gurunathreddy.m@gehealthcare.com, dattesh.shanbhag@gehealthcare.com, deepa.anand1@gehealthCare.com).}}

\maketitle

\begin{abstract}
The high cost of obtaining accurate annotations for image segmentation and localization makes the use of one and few shot algorithms attractive. Several state-of-the-art methods for few-shot segmentation have emerged, including text-based prompting for the task but suffer from sub-optimal performance for medical images. Leveraging sub-pixel level features of existing Vision Transformer (ViT) based foundation models for identifying similar region of interest (RoI) based on a single template image have been shown to be very effective for one shot segmentation and localization in medical images across modalities. However, such methods rely on assumption that template image and test image are well matched and simple correlation is sufficient to obtain correspondences. In practice, however such an approach can fail to generalize in clinical data due to patient pose changes,  inter-protocol variations even within a single modality or extend to 3D data using single template image. Moreover, for multi-label tasks, the RoI identification has to be performed sequentially. In this work, we propose foundation model (FM) based adapters for single label, multi-label localization and segmentation to address these concerns. We demonstrate the efficacy of the proposed method for multiple segmentation and localization tasks for both 2D and 3D data as we well as clinical data with different poses and evaluate against the state of the art few shot segmentation methods.
\end{abstract}


\section{Introduction}
\label{sec:introduction}
\IEEEPARstart{M}{ost} of radiology tasks necessitate localization of anatomy or landmark, and lesion, which is laborious, repetitive and prone to errors, especially due to variations in patient pose, conditions, and disease. Advances in Artificial Intelligence(AI) has simplified the effort by utilizing various deep learning paradigms (supervised training, transfer learning, active learning) to automate the tasks for localization and landmarking. Most of these approaches are data driven and the burden and associated risks have shifted from end-application user to AI developer for getting data manually annotated for further usage. 

There is emerging body of work on performing few shot segmentation aiming to train Deep Learning(DL) network with only few annotated datasets. Recent advancements in the ability of ViT based and Diffusion based models \cite{Zhang:arXiv:2023} for obtaining point correspondences have shown great promise in leveraging pixel level similarity for region correspondences between pairs of images \cite{Hamilton:arXiv:2023}. A recent work \cite{Anand:arXiv:2023} demonstrates the utility of using such feature correspondences for localization of landmarks in medical images as well as obtaining organ segmentation by chaining it with SAM \cite{Zhang:arXiv:2023:Personalize}. 

However, most of these one-shot methods are based on 2D images and expect that the template or exemplar image are well matched to the new test image for correspondence matching. In our experience, such methods do not scale well for 3D imaging volume nor in cases where image representation between exemplars and test data are mismatched due to geometric changes, protocol changes or disease conditions. To overcome these limitations, in this paper, we propose to leverage sub-image level features derived from fine-tuned foundation models and by training a lightweight task adapter, using the derived pixel-features for obtaining multi-label localization which generalizes well across these challenges.

\section{Related Work}
Methods exploring the use of self-supervised learning and transfer learning techniques for few shot learning have been shown to be effective. For instance, the paper \cite{Qi:Article:2023} introduces double RoI heads into the existing Fast-RCNN to learn more specific features for novel classes. It also uses self-supervised learning to learn more structural and semantic information. 

Specialized network architectures play a crucial role in few-shot segmentation by effectively learning from limited data and improving the model’s performance. Prototype networks have been shown to be superior for few-shot learning scenarios \cite{Zhu:Article:2023} \cite{Wang:Proceedings:2019}, using support and query image paradigm, similar to our approach. The key idea is to decompose the holistic class representation into a set of part-aware prototypes, capable of capturing diverse and fine-grained object features. These networks often leverage large set of unlabeled data to enrich the part-aware prototypes \cite{Liu:Proceedings:2020}, which though possible in natural images may be impractical for data-sparse and diverse situations - common in medical imaging domain.

Methods that use text prompts for image segmentation/localization \cite{Wang:arXiv:2023}, \cite{Liu:arXiv:2023} are gaining significance area in computer vision. They leverage the power of natural language to guide the task allowing flexibility, interactivity and the potential to handle a wider range of tasks within the scope provided as part of supervised training. Consequently, these methods do not scale well for out-of-domains images and pointed landmarks identification tasks.

Recent works take advantage of the capability of ViT and diffusion models to extract patch level features \cite{Zhang:arXiv:2023} and leverage these to pixel correspondence tasks and extend these for localization  and segmentation tasks by chaining them with other foundation models to obtain few shot leaning capability \cite{Anand:arXiv:2023}. A similar technique \cite{Zhang:arXiv:2023:Personalize} leverages latent sub-pixel level features Segment Anything Model (SAM \cite{Kirillov:arXiv:2023}) and train a small Adapter for a single template image in order to obtain segmentation regions on other similar images.

Though the above listed methods are effective for 2D image tasks based on natural images, they do not extend naturally to 3D medical volumes with geometric, protocol and patient variations. Moreover, these methods are based on pixel feature extraction and simple similarity metrics thus requiring user to rely on task-dependent heuristics which need to be manually tuned.

\section{Method}
\begin{figure*}[h]
	\centering
	\includegraphics[width=1\textwidth]{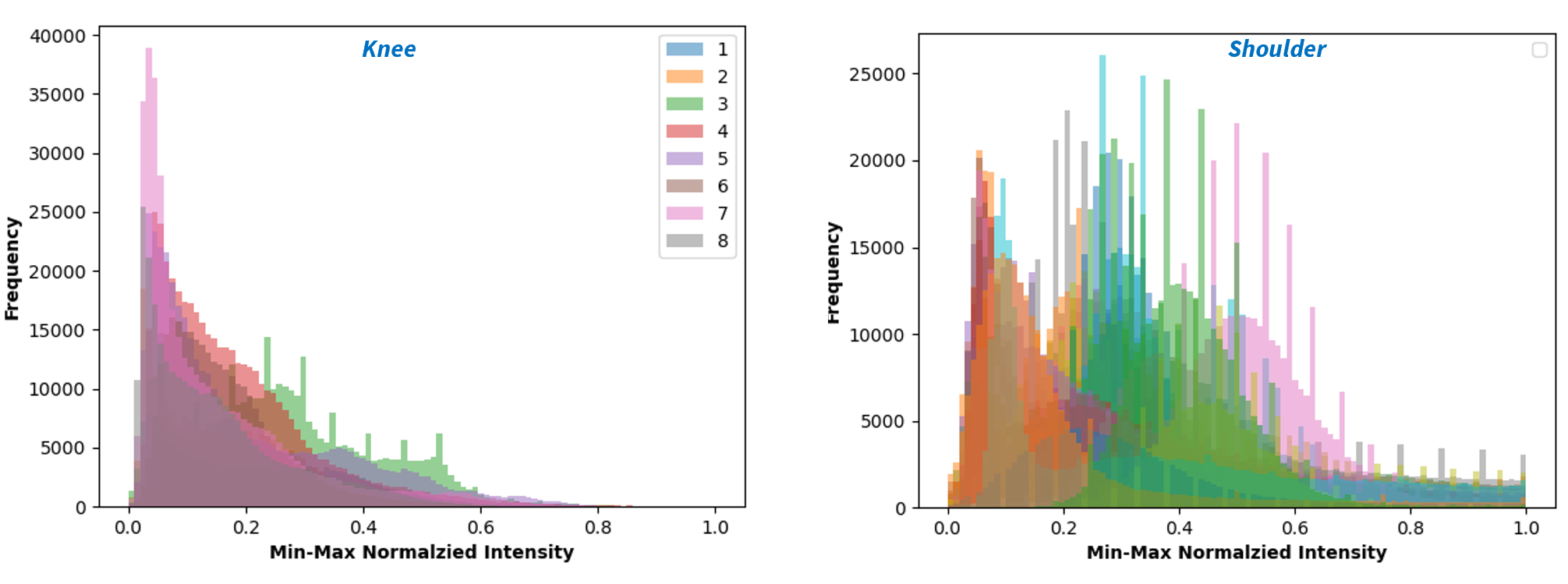}
	\caption{The differences in image intensity distributions of a cohort of MRI data for knee and shoulder is shown here. We observe heterogeneity in distributions across cases which will necessitate manual threshold adaptation, which is overcome with our proposed approach.}
	\label{fig:histogram}
\end{figure*}

Previous work \cite{Zhang:arXiv:2023}, \cite{Anand:arXiv:2023} on few-shot segmentation or localization have relied on template and target images being similar in structure and contrast as captured in feature semantics. To segment similar regions as marked in template, the features obtained at pixel level from ViT models are correlated using similarity metric and further thresholded to localize the region of interest \cite{steck2024cosine}. However, obtaining this threshold is not trivial due to large variations in image intensity (including across slices) (See Fig. \ref{fig:histogram}) and contrast (due to site based intra-protocol changes) observed in imaging data in clinical practice. This impacts the segmentation results in multiple ways: \textbf{a)} For  given imaging volume in a subject, multiple templates have to be chosen to get reliable segmentation to accommodate intensity changes across slices and \textbf{b)} Inability to scale segmentation reliably across the diverse pool of imaging data due to various patient conditions or site specific imaging protocols. The need for manual tuning hinders practical adoption of template based few shot region segmentation and a method is acutely needed to overcome this. 

In this work, we designed a contrastive learning based adapter which is trained to  automatically determine which of the pixel level feature vectors in target image are similar to the template pixel level feature vector for given set of ground-truth label(s); thereby obviating the need for user to fix any threshold. We hypothesize that the proposed contrastive adapter is better at discriminating learnt representation even when intensity appearance/ViT features are almost homogeneous.

The main contribution of our work is to utilize the features provided by foundation models (fine-tuned DINOv2) to enable accurate segmentation/localization using one or few template images. To enable this in clinical settings we do the following a) train a Siamese architecture based contrastive learning adapter to learn robust feature matching metric b) ability to enable multi distinct region localization simultaneously using a single model, c) Ability to generate automatic prompts for SAM model to obtain refined segmentation and d) by virtue of this, ability to annotate 3D volumes with minimal templates.

Transformer architectures enable deriving patch level features for images, which can be interpolated to pixel level features \cite{Anand:arXiv:2023}. Next, we utilize these pixel level features to correlate the region marked in template image with similar region in target image. 

\textbf{Basic adapter}: A straightforward approach to obtain this would be to correlate template image pixel feature vector with the feature vectors for each of the pixels in the target image using cosine similarity measure and threshold the correlation map at a fixed heuristic threshold (= 0.5) to obtain the binary localization mask. This is referred to as basic adapter and involves no training. The basic adapter is not well suited to account for variations in clinical data or multi-label tasks and hence we design two more sophisticated trained adapters to handle these limitations, as described below.

\textbf{Classification adapter}: 
The Classification adapter is a lightweight feed-forward three layer neural network, taking as input the pixel features and trained with a cross-entropy loss function. During inference, pixel level features for all pixels in the target image are fed to the model to obtain the pixel-wise binary or multi-label localization mask. The localization mask is subject to connected component analysis to remove any stray/isolated pixels.We find that following this simple classification approach leads to sub-optimal results (especially in MRI data) as discussed in the experimental section.

\begin{figure*}[h]
	\centering
	\includegraphics[width=0.6\textwidth]{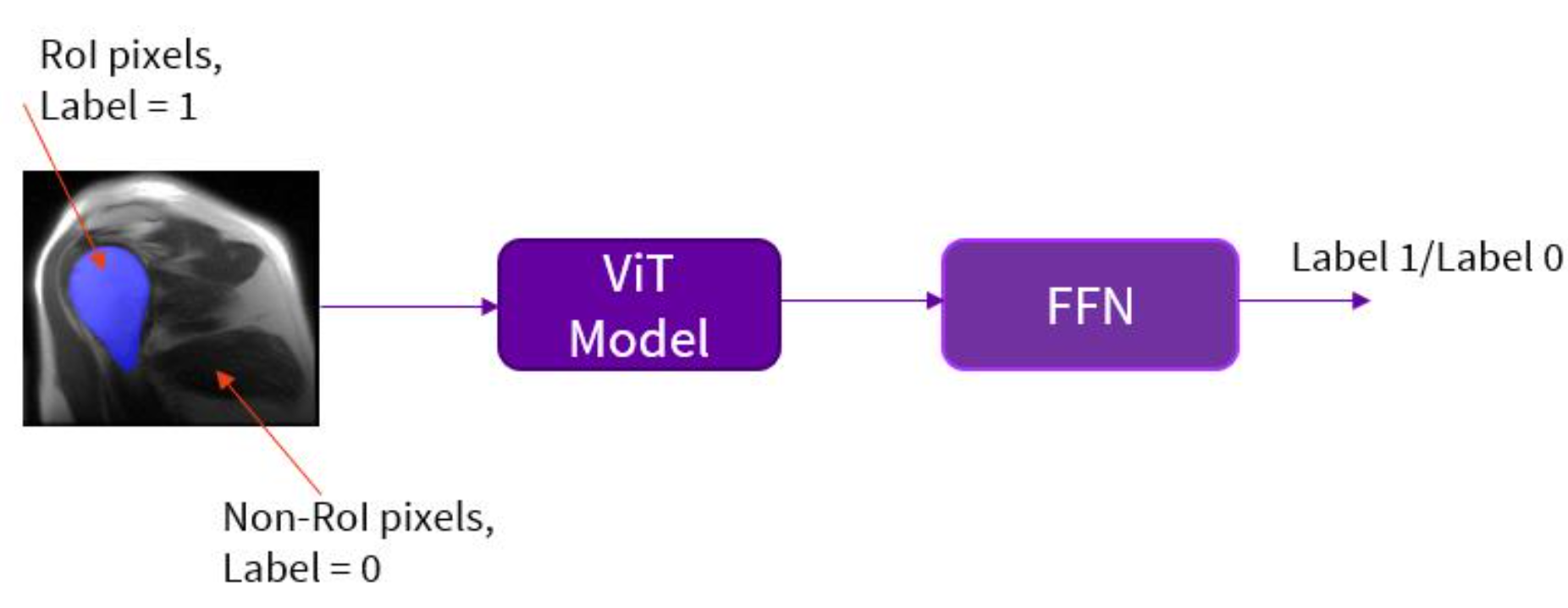}
	\caption{The pixels within the RoI are considered as label 1 and outside as label 0. A simple classifier is trained to predict the labels of the pixels from the feature vectors, derived from the trained DINOv2 model. 
	}
	\label{fig:classadapt}
\end{figure*}

We have trained both binary and multi-label localization classifiers as described below:
a) Binary localization: We  train a binary classifier to label pixels within an ROI as foreground and randomly selected pixels outside the ROI (matching the number of pixels in foreground to maintain class balance) as background shown in Fig.~\ref{fig:classadapt}.  
b) Multi-label localization: We define a background as set of pixels not belonging to any of the multi-label ROI. We then train a multi-class classifier to label pixels belonging to different ROI labels and background (i.e number of classes = number of labels+1).  

\begin{figure*}[h]
	\centering
	\includegraphics[width=1\textwidth]{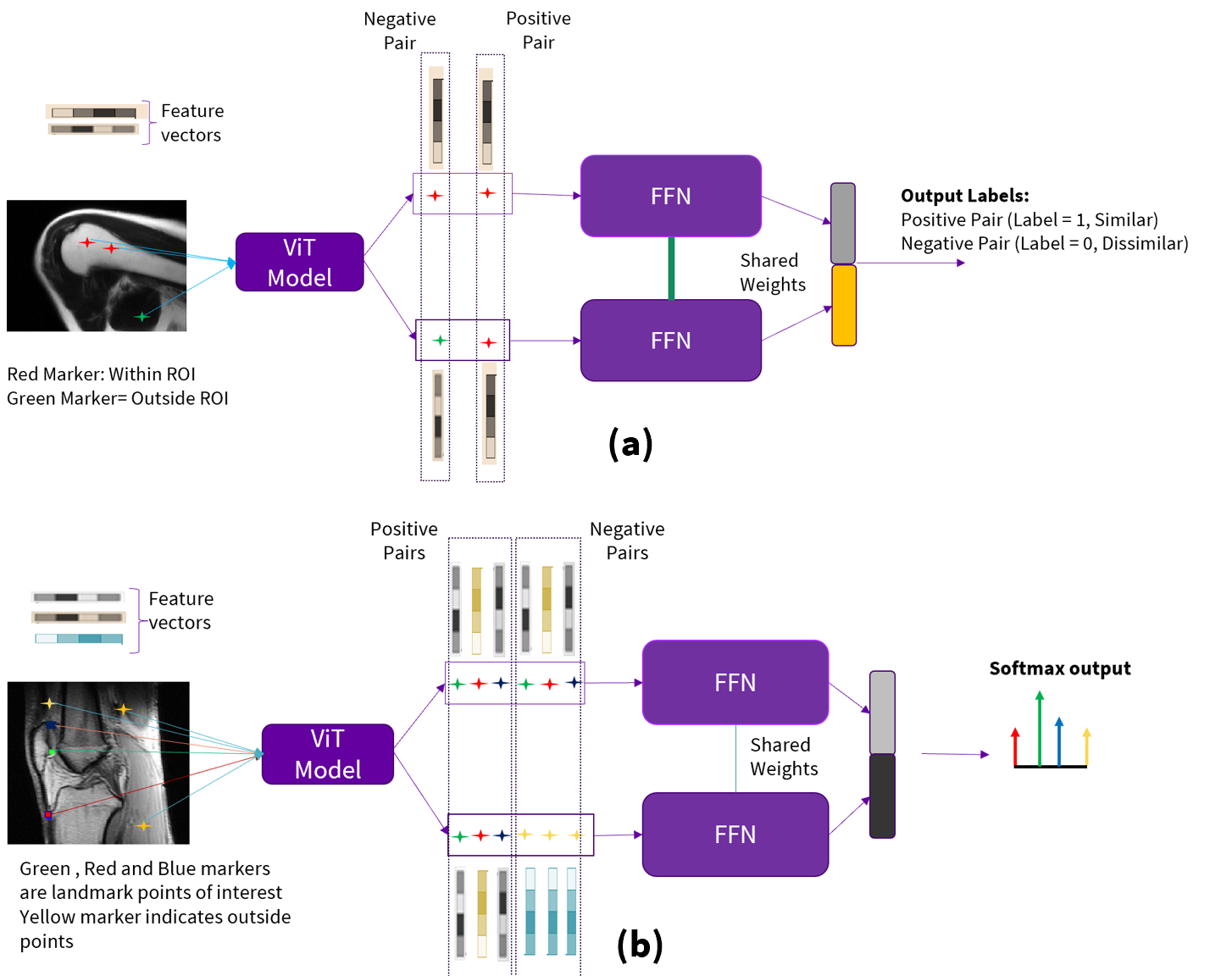}
	\caption{Contrastive similarity model for binary label localization (a): The contrastive model consists of two subnetworks. Pairs of positive pixels (red markers) are chosen from the RoI (shoulder tibia) and negative pairs - one from RoI and the other from outside the RoI (green marker). Feature vectors for pairs of pixels are derived from the finetuned DINOv2 ViT model. Feature vector pairs are passed to the network to learn the similarity measure for localization by minimizing the distance between the positive pairs and maximizing the distance between the negative pairs. The model is trained with cross-entropy loss function to obtain localization map to alleviate the thresholding. (b) Extension of proposed approach for multi-label localization: Multiple -labels for knee localization are shown here: TT (red box), patella (green box), and UPT (blue box). For each landmark paired positive and negative pixel pairs are sampled. The contrastive model is trained with these paired pixel features using cross-entropy loss.}
	\label{fig:Contrastive}
\end{figure*}

\textbf{Contrastive adapter}: We follow the Siamese architecture with a contrastive loss (cross-entropy loss) as described in Eq [8] from \cite{wang2022medclip}. The network design is shown in Fig. \ref{fig:Contrastive}(a) for binary localization  and in Fig \ref{fig:Contrastive}(b) for multi-label localization. It consists of two fully connected lightweight feed-forward neural network (Fig. \ref{fig:Contrastive}) with shared weights. The outputs of the two sub-networks are concatenated and fed to the softmax layer.

\begin{figure*}[h]
	\centering
	\includegraphics[width=0.2\textwidth]{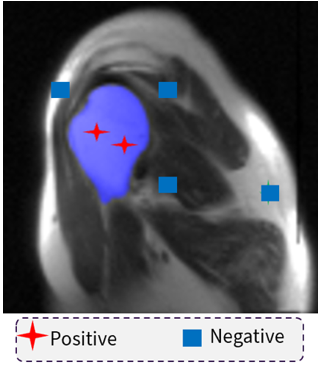}
	\caption{We choose pairs of pixels from the RoI (red markers) as positive pair pixels. Whereas we choose a pixel from the RoI (red) and a pixel outside the RoI (blue) and pair them as negative pixels for contrastive learning. In our experiments, we have chosen the negative pixel 10 pixel away from the RoI but in our further experiments we found that choosing negative pixels anywhere in the non-RoI region do not hamper the results.}
	\label{fig:posnegpix}
\end{figure*}

a) \textbf{Binary localization}: We consider pairs of pixels to be positive if both belong to the same RoI. We create negative pair pixels by choosing one pixel from RoI and another random pixel from outside the RoI as shown in Fig.~\ref{fig:posnegpix}. Feature vectors corresponding to pairs of positive and negative pixels are passed through the sub-networks, as shown in the Fig. \ref{fig:Contrastive} (a), to predict how similar the input pair feature vectors are with each other. This is achieved by minimizing the distance between positive pairs and maximizing distance between negative pairs. We train the model with cross-entropy loss function to obtain the binary mask as localization map.

b) \textbf{Multi-label localization} (Fig. \ref{fig:Contrastive} (b)): For a given label RoI,  we consider pairs of pixels to be positive if both belong to the same  RoI. Negative pairs are created by choosing one-pixel in chosen label ROI and one outside the chosen label ROI. This process is repeated for all the labels in consideration. We train the contrastive model with these paired pixel features using contrastive loss to obtain the softmax probability distribution. The index of the softmax distribution with highest probability is associated with the corresponding label.         

\begin{figure*}[h]
	\centering
	\includegraphics[width=1\textwidth]{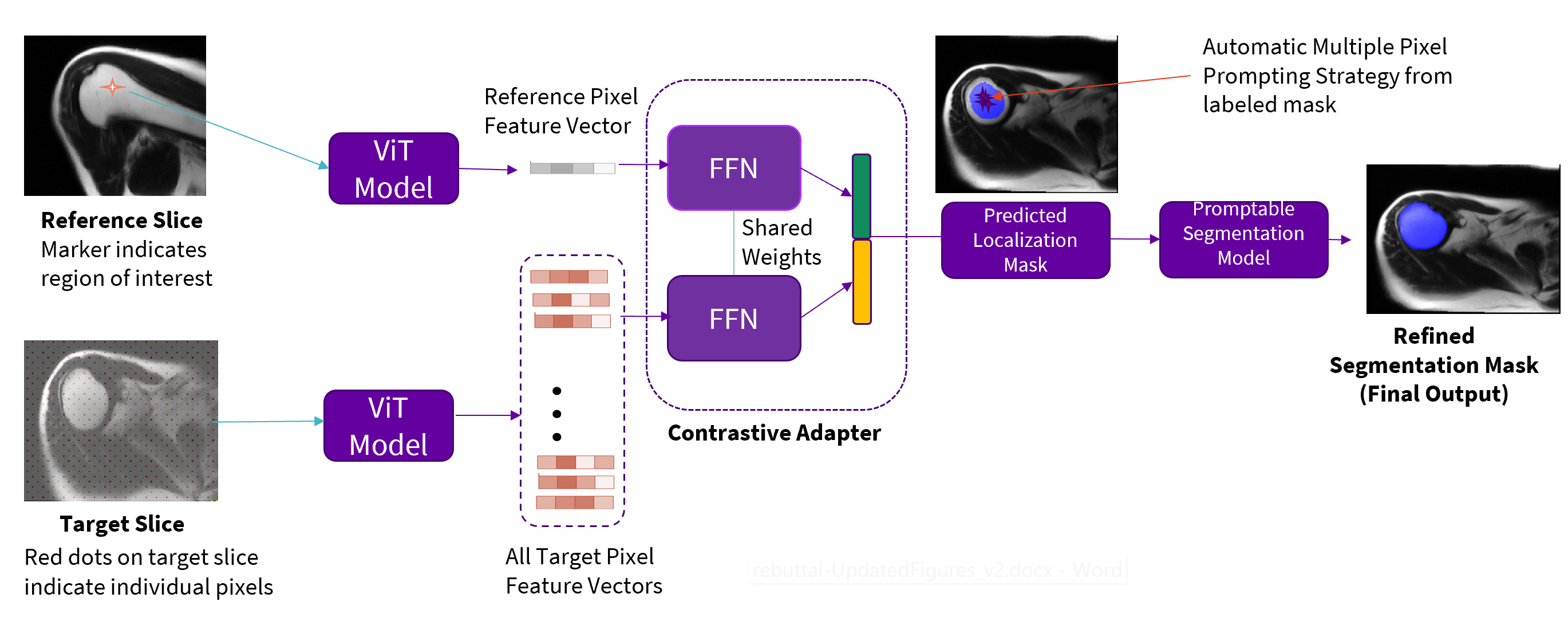}
	\caption{Inference procedure for localization and segmentation: Feature vectors for pixels from RoI in template image  are derived using ViT and treated as template/reference pixel feature vectors. Similarly, feature vectors for all pixels in target image are computed. Reference and target feature vectors are paired and given as input to contrastive model to obtain localization region. The output undergoes a connected component analysis to remove stray/isolated pixels. From the localized region, ten pixels are randomly chosen and used as prompts to SAM for refined segmentation}
	\label{fig:ContrastiveInference}
\end{figure*}

During inference for Contrastive adapter, as shown in Fig. \ref{fig:ContrastiveInference}, we choose reference pixels from each label RoI and compute the feature vectors and treat them as template feature vectors for that RoI. Similarly, for the target slice, we compute the feature vectors for all the pixels. We pair each template feature vector with all the feature vectors from the target image and pass the pairs to the appropriate contrastive model (binary or multi-label), which then assigns label(s) to each pixel in target image according to the highest probability value from the softmax distribution. Note from Fig. \ref{fig:ContrastiveInference} that reference pixel feature vector is fed to the upper Feed Forward Neural network (FFN) while target pixel feature vector is fed to the lower FFN. The localization mask is subject to connected component analysis to remove any stray/isolated pixels. If the task is landmark localization, this is the final output. If the task is segmentation of structure, we randomly choose ten pixels from localized region and prompt the SAM to refine the segmentation of the object of interest. The output of SAM is the final segmentation task output. 

\begin{figure*}[h]
	\centering
	\includegraphics[width=1\textwidth]{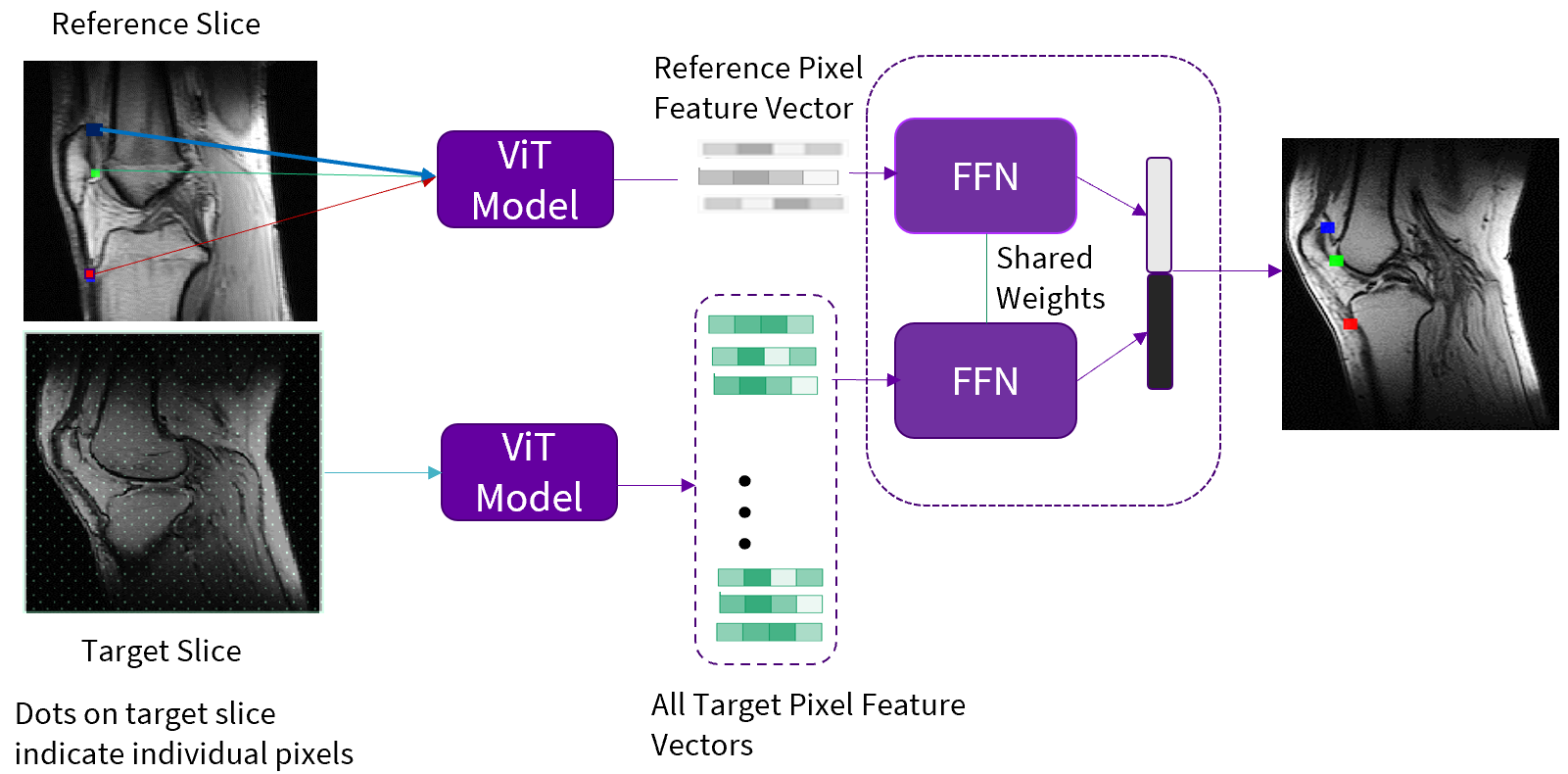}
	\caption{Multi-label localization inference: During inference, each pixel feature in the target image along with the features from the RoI from source are passed to the adapter. The result is the probability distribution over all the labels - the highest probable label is chosen as the label per pixel.}
	\label{fig:multilabellocInf}
\end{figure*}

The simultaneous multi-label localization using a single model using the proposed contrastive method is shown in the Fig. \ref{fig:multilabellocInf}, using a single adapter by comparison with pixel from RoI in source image.

\section{Experiments and Results}
We evaluate the efficacy of the proposed few-shot method for five different medical 3D image tasks, two of them for segmentation and three for localization. To benchmark the performance, we compare the results with recently proposed and popular few shot methods such as UniverSeg \cite{Butoi:arXiv:2023} and PerSAM \cite{Zhang:arXiv:2023:Personalize}. Both of these methods enable users to tackle new segmentation task without the need to train or fine-tune a model, adapting to a new segmentation task at inference based on few sample examples.

For PerSAM and basic adapter based localization methods (which are designed for binary tasks), for multi-label tasks, we aggregate the predictions using each label in the template pool.

\subsection{Datasets \& Metrics}

\textbf{DINOv2 training}: DINOv2 model was trained on a pool of 12000 MRI image volumes, acquired in-house on a range of MRI scanners (1.5T, 3T) and across anatomies. This model was used to extract pixel-level features for the proposed methods described in the paper. 

\textbf{Proposed Experiments}: We evaluate the trained different customized adopters on five datasets. 
a) \textbf{Shoulder segmentation} on MRI 3-plane stack (in house): A set of 43 image volumes comprising of axial, coronal and sagittal orientations are used for segmentation task. For few shot prediction using the proposed approach and parallel methods, five volumes were used for training the binary adapters. b) \textbf {CT Liver} 3D segmentation \cite{Wasserthal:arXiv:2022}: 10 volumes chosen from TotalSegmentator dataset were used along with their liver mask for evaluation. All axial images containing liver from five CT volumes were used for training the binary adapters. c) \textbf {Whole Body} MRI Localizer Images (in house): A set of five whole-body image volumes were evaluated for identification of three different landmark points across slices: i) Wholebody interface between lung and neck (WILN) ii)  inferior to costophrenic angle (WICA) iii) top of illiac crest (WIC). Four volumes were used for training the binary adapters
d) \textbf {MRI Knee} Sagittal multi-label Localization (clinical site data): A set of 8 sagittal knee volumes were evaluated for localization of three specific landmarks: i) Knee Patella Base(KPB), ii) Knee Apex Lateral (KAL), iii) Knee patellar ligament insertion point (KPLIP). This is a task where multiple types of landmarks were detected and hence as a multi-label contrastive learning approach was adopted. Seven slices containing the landmarks were used for training the multi-label adapters. e) \textbf {CT Headneck} \cite{Podobnik:Article:2023}: We perform localization of the i) pairs of eyes (HNEye), ii) pairs of optic nerve (HNON) and iii) optic chiasm (HNOC) on CT head-neck 3D image volumes. Five axial CT images were used for training multi-label adapters. 10 CT axial volumes were used for testing landmark localization.
\subsection{Details on Training and Inference}


The above tasks have been chosen to demonstrate fairly good generalization of approach across modalities (MR \& CT) and tasks (segmentation \& localization). For the in-house datasets the ground-truth (GT) was generated by a trained radiologist. To evaluate performance for segmentation tasks, we compute the intersection over union (IoU) measure \cite{van2019deep} between GT and predictions and for localization task we use localization accuracy Eq. \cite{kang2021accurate} - which measures the ratio of cases for which the euclidean distance between the ground truth and the predicted landmarks are below ten pixels.

\begin{table*}[h]
	\begin{center}
		\resizebox{16cm}{!}{%
			\begin{tabular}{| l | l | l | l | l | l | l | l | l | l | }
				\hline
				\textbf{Method} & \textbf{KPB} & \textbf{KAL} & \textbf{KPLIP} & \textbf{WILN} & \textbf{WICA} & \textbf{WIC} & \textbf{HNEye} & \textbf{HNON} & \textbf{HNOC} \\ \hline
				\textbf{Universeg} & 0.75 & 0.375 & 0.521 & 0.71 & \textbf{0.462} & 0.265 & 0.728 &0.198 &0.35 \\ \hline
				\textbf{PerSAM} & 0.21 & 0.248 & 0.276  & 0.319 & 0.358 & 0.212 & 0.54 & 0.34 & 0.31 \\ \hline
				\textbf{Basic Adapter} & 0.5 & 0 & 0 & 0.71 & 0.312 & 0.422 & 0.212 & 0.310 & 0.127 \\ \hline
				\textbf{Classification Adapter} & 0.25 & 0.375 & 0.375 & 0.562 & 0.187 & 0.322 & 0.761 & \textbf{0.811} & 0.684\\ \hline
				\textbf{Contrastive Adapter} & \textbf{0.88} & \textbf{0.88} & \textbf{0.77} & \textbf{0.812} & 0.45 & \textbf {0.733} & \textbf{0.813} & 0.534 & \textbf{0.631} \\ \hline
			\end{tabular}
		}
	\end{center}
	\caption{Comparison of localization accuracy for MR Knee, MR WholeBody (WB) and CT HeadNeck (HN) landmarks detection tasks via different methods.}
	\label{table:results}
\end{table*}

\subsection{Results}

\begin{figure*}[h]
	\centering
	\includegraphics[width=1\textwidth]{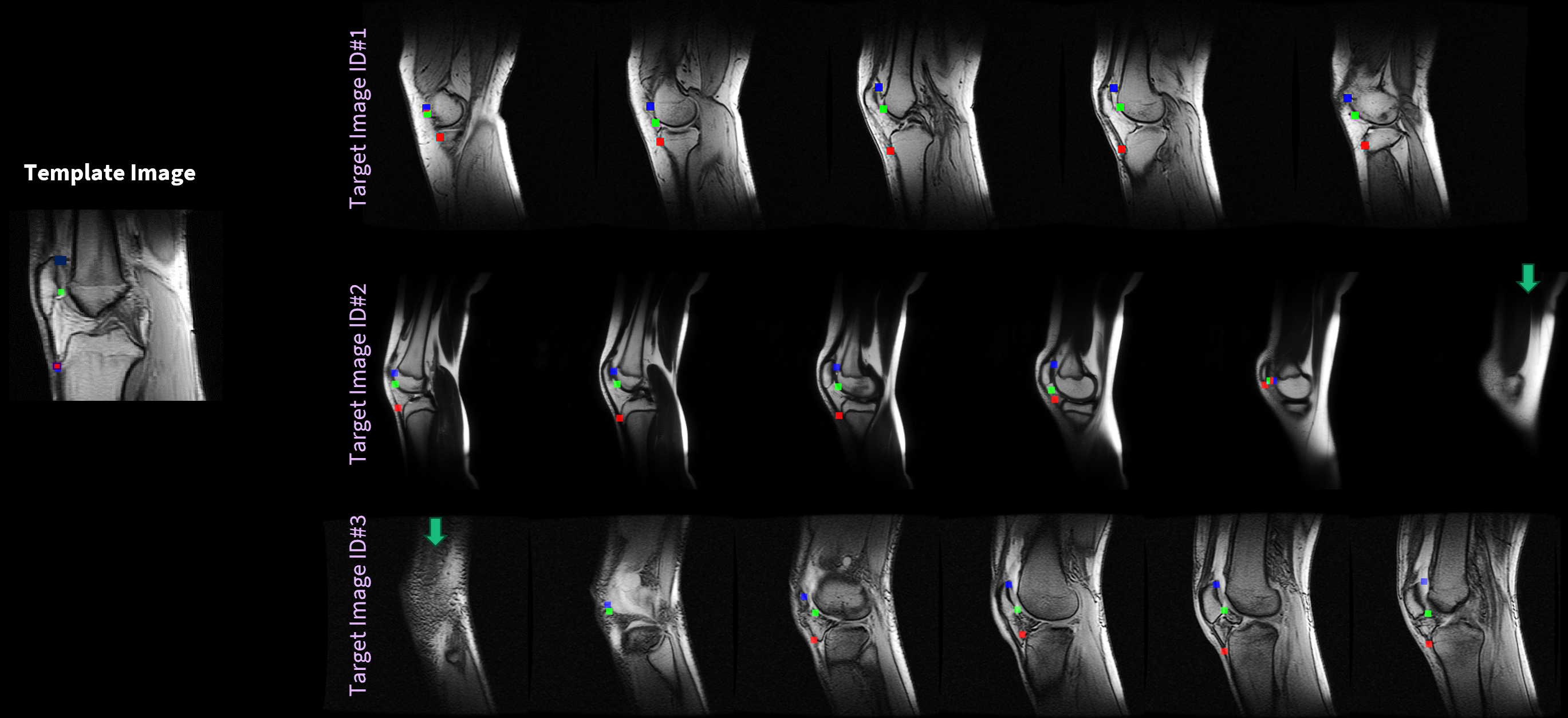}
	\caption{Output of localization of the 3 knee landmark points obtained from contrastive multi-label adapter based on landmarks on template image (left panel). Notice excellent localization of landmarks across knee slices and capability to prevent localization on slices which do not contain knee anatomy (green arrows). All the landmarks are predicted in single shot by the proposed method.}
	\label{fig:result2}
\end{figure*}

\begin{figure*}[h]
	\centering
	\includegraphics[width=1\textwidth]{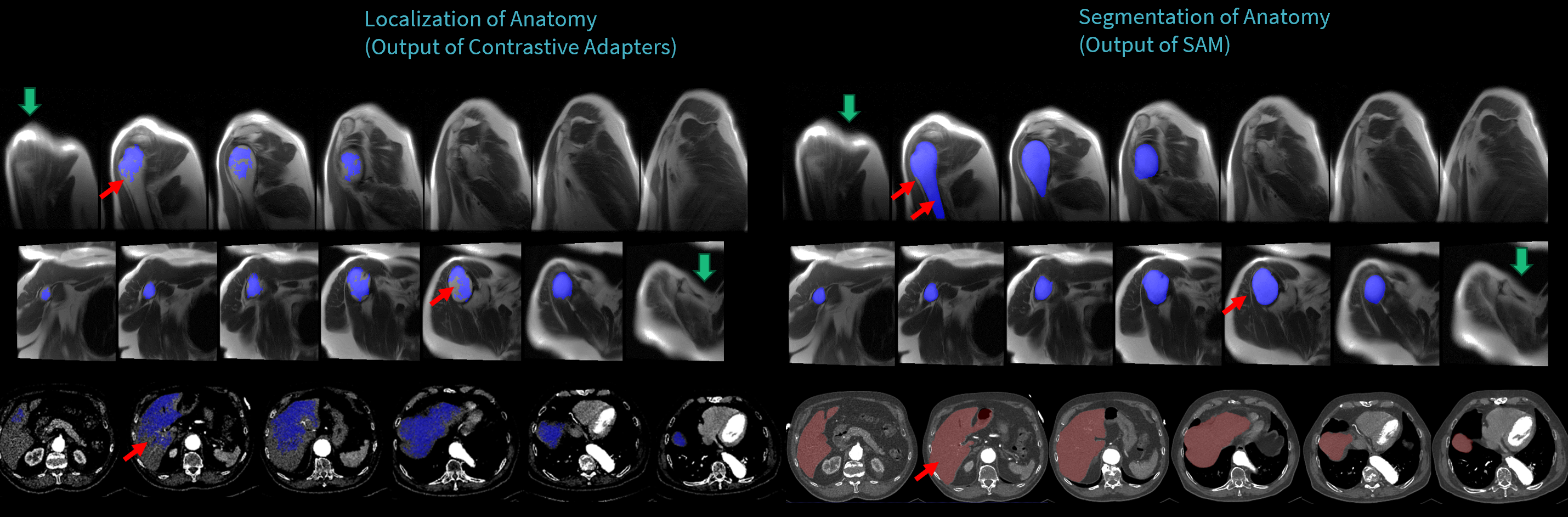}
	\caption{The left panel shows the output of localization obtained from contrastive adapter. We notice excellent localization capability across slices and orientations (1st row = sagittal and 2nd row = axial), including true negative capabilities (not localizing regions which are not part of template for that anatomy) as indicated by green arrows. However as indicated by red arrows, the localization does not cover the entire anatomy. Therefore, we chain the output of localization mask as prompts to SAM model, which generates complete anatomy segmentation.}
	\label{fig:result3}
\end{figure*}

The qualitative visualization of the proposed {\it contrastive adapter} for multiple-label localization and segmentation is shown in Figs \ref{fig:result2} and \ref{fig:result3}, respectively. We notice that there are no false positives on slices which do not contain the anatomy of interest in the image stack (shown using green arrows) and the localization are sharp and constrained within the RoI. 

\begin{figure*}[h]
	\centering
	\includegraphics[width=1\textwidth]{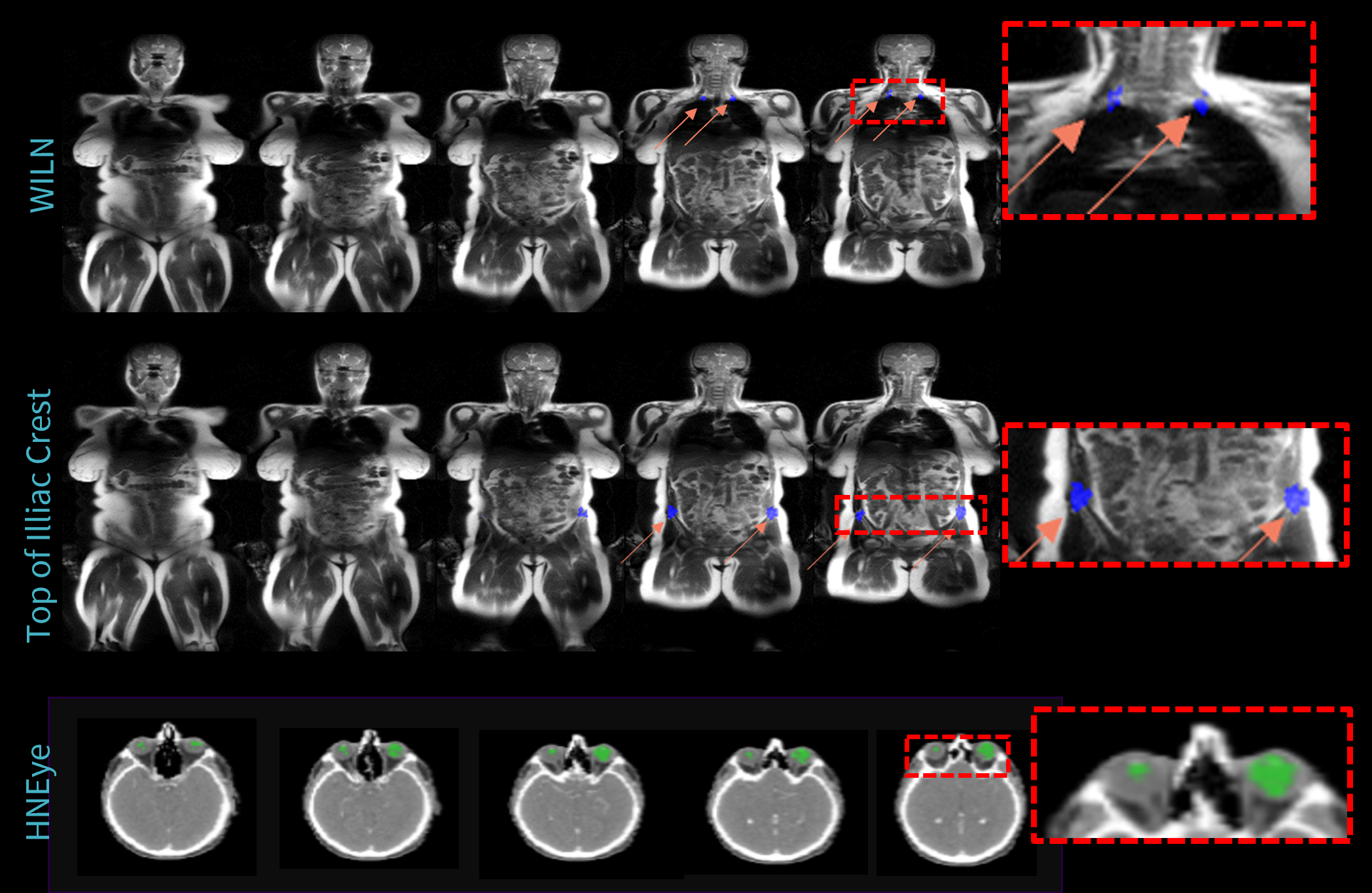}
	\caption{Localization outputs for wholebody(WIC,WILN) and CT Headneck(HNEye) using the proposed contrastive adapter}
	\label{fig:result1}
\end{figure*}

Fig. \ref{fig:result1} shows results of localization of different landmarks for wholebody and CT Headneck anatomies and accurate localization results are observed in all cases.

\begin{figure*}[h]
	\centering
	\includegraphics[width=1\textwidth]{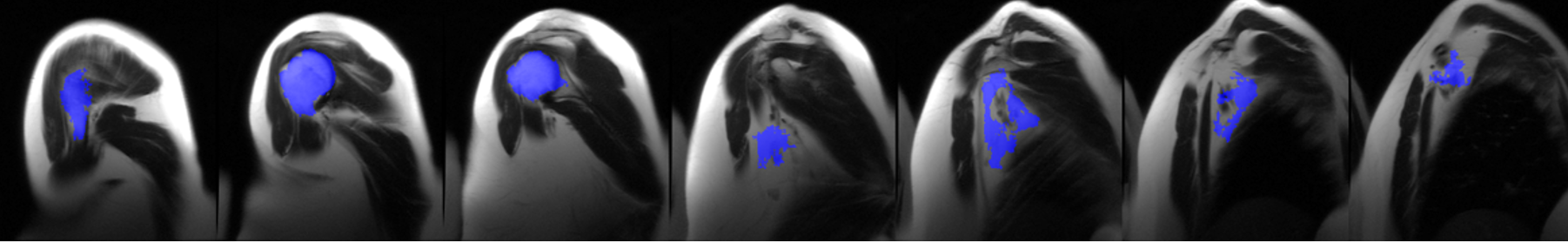}
	\caption{Localization masks obtained by the classification adapter for shoulder localization on the entire shoulder volume is shown in this figure. We observe that the classification adapter results in false positives by localizing on the slices where shoulder bone is not present (slice 1, 4, 5, 6, 7). Further, prompting the SAM with the wrong pixels from the false positive pixels results in wrong segmentation.}
	\label{fig:classifierShoulder}
\end{figure*}

We show the differences in the qualitative results obtained using basic adapter and classification adapter as compared to the contrastive adapter, using the MRI shoulder as an example. Considering 3D volumes, where the appearance of a desired RoI can vary across slices and taper towards the ends, Fig. \ref{fig:classifierShoulder} shows the localization masks obtained using the classification adapter on the entire shoulder volume. Multiple false positives on slices are observed as described in the Fig. \ref{fig:classifierShoulder}.

\begin{figure*}[h]
	\centering
	\includegraphics[width=1\textwidth]{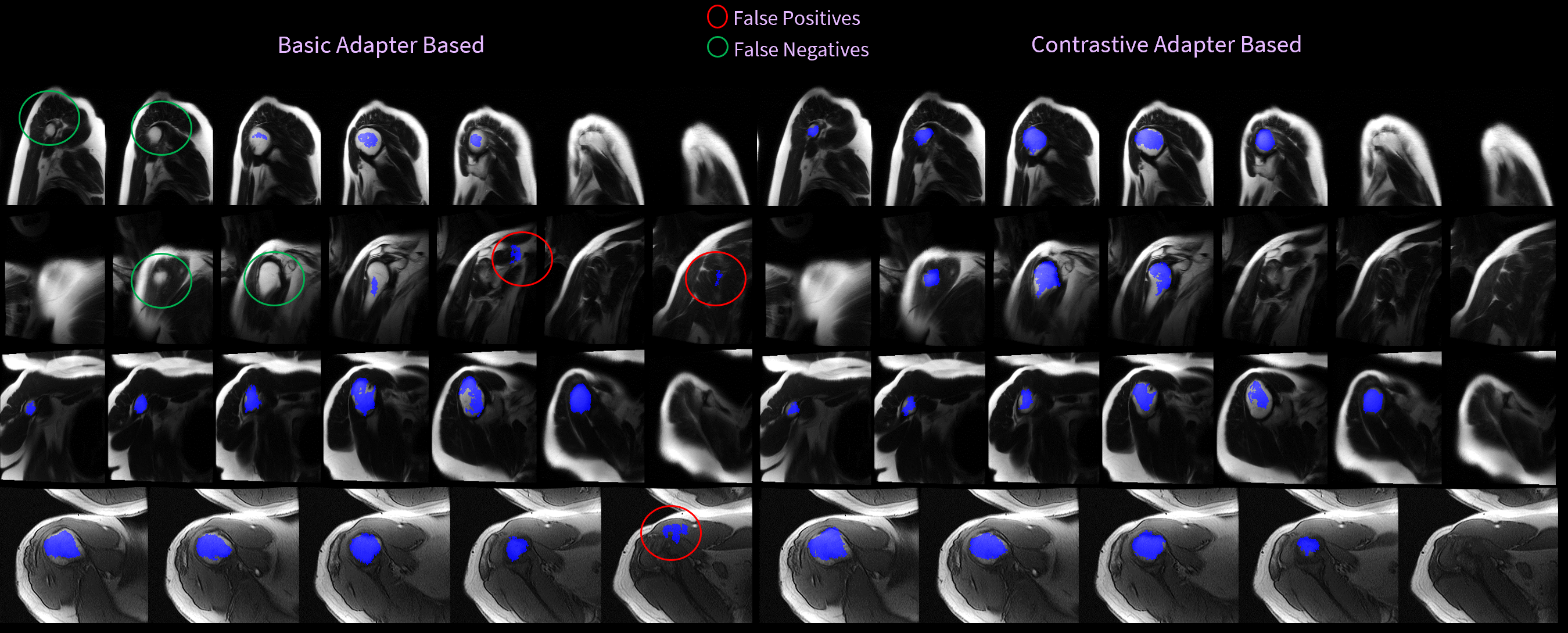}
	\caption{Qualitative comparison between simple cosine similarity measure and proposed contrastive method is shown here for MR Shoulder. The cosine similarity measure for localization results in false positives and false negatives as highlighted. The proposed contrastive method, trained to contrast between positive and negative pair of pixels results in improved localization quality.}
	\label{fig:similarityvscontrastive}
\end{figure*}

The qualitative comparison between localization obtained through basic adapter and the contrastive adapter localization is shown in the Fig. \ref{fig:similarityvscontrastive}. We can see the vast improvement in avoiding false positive and negative using contrastive adapter  as compared to basic adapter.

\begin{figure*}[h]
	\centering
	\includegraphics[width=1\textwidth]{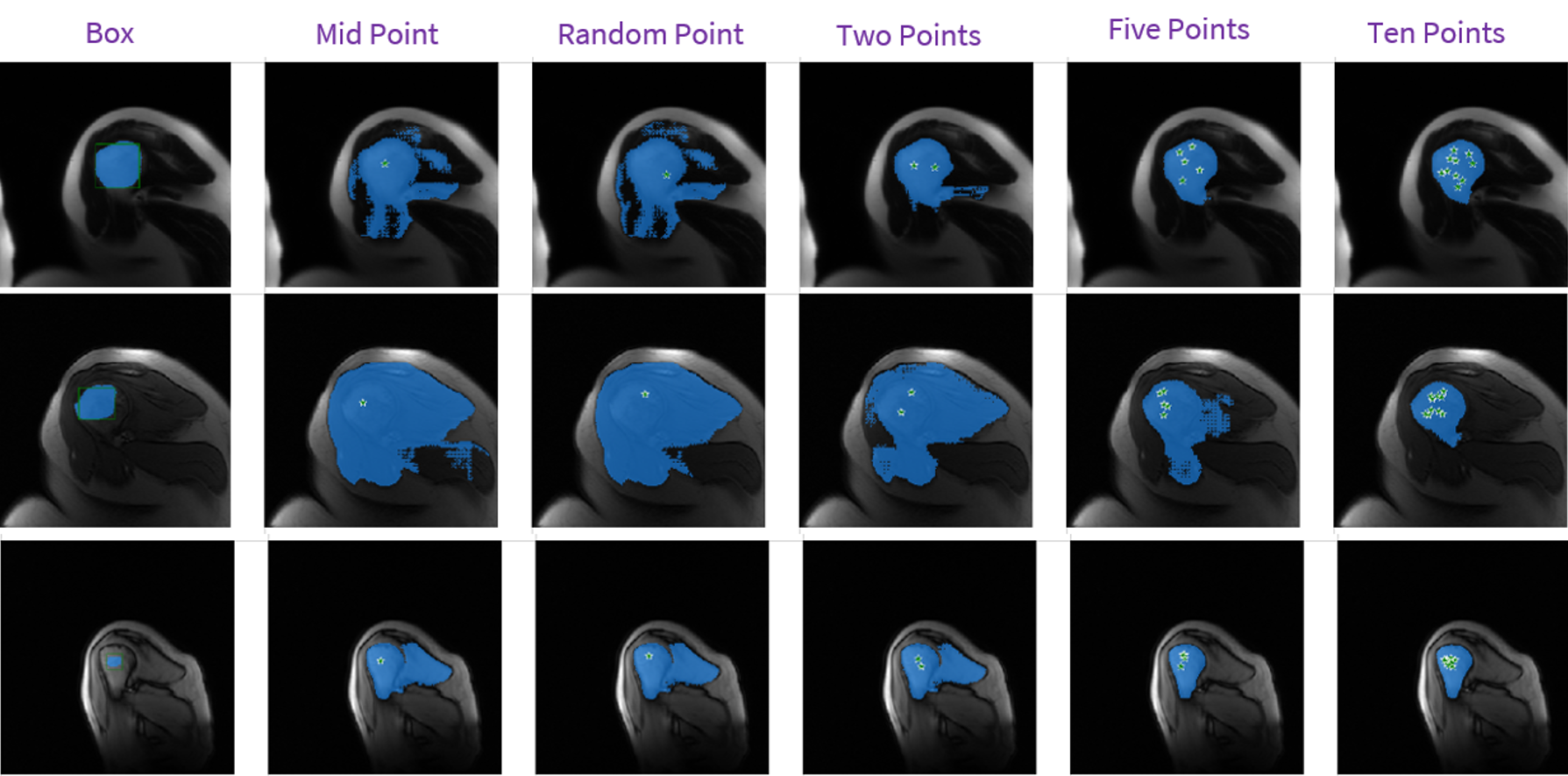}
	\caption{Comparison of multiple strategies adopted for prompting SAM after obtaining localization. We can see that the SAM prompted by box prompt created by creating a box around the predicted localization results in poor segmentation compared to randomly choosing ten pixels as prompts with in the automatically localized regions. Empirically we found that ten pixels are more than sufficient to obtain the acceptable segmentation.}
	\label{fig:promptingSAM}
\end{figure*}

We explored several prompting strategies for SAM from the RoI obtained through the proposed approach. Fig. \ref{fig:promptingSAM} shows a quantitative comparison of the different strategies. It was found that 10 randomly chosen pixels from localized region prompts was the optimal in terms of quality of results and was adopted for all our experiments.

Table \ref{table:results} shows the results of evaluation of the different approaches for localization tasks, for nine different landmarks across multiple anatomies and modalities. Localization accuracy is a measure of average binary agreement for the given predicted landmark to be within 10 pixels of GT pixels. In most of the landmarks, we notice that our proposed contrastive adapter outperforms other methods except for WICA, where it is marginally lower than Universeg. We notice that basic adapter performs poorly since it is based primarily on simple cosine distance-based similarity metric which is thresholded at a fixed value (=0.5). Changing the threshold can improve the accuracy, but also results in large false-positives and is automatically discarded by the post-processing scheme. Classification adapter performs reliably in CT datasets which have standardized intensity values and hence therefore less impact on feature variability. However, in MRI datasets, the intensity variations are large (as seen in Fig \ref{fig:histogram}) and consequently we notice poor performance in all the MRI tasks. We reason that this is due to inability of classification adapter to impose similarity matching condition for pixels belonging to same ROI. The contrastive adapter on virtue of being able to cluster semantically similar features vectors results in robust performance in most of the cases; especially in MRI datasets. 

Table \ref{table:SegAccuracy} presents the results for segmentation tasks. As is evident  from the IoU metric, the proposed method performs the best for both the tasks across anatomies and modalities (MR, multi-orientation for shoulder segmentation and CT-axial for liver segmentation). This is due to ability of our proposed contrastive adapter to leverage the localization capabilities of ViT and learn the  ROI similarities for a given task by pulling the similar pixel feature vectors closer despite intensity variations in images and pushing the non-ROI pixel feature vectors away; despite similarities of regions in intensity space (See Fig. \ref{fig:similarityvscontrastive} ). Without the contrastive adapter, we notice that the localization capabilities of ViT alone cannot reliably avoid false positives and negatives due to overlap of intensity or contrast of structures (e.g. fat in bone marrow vs fat in skin layer) (See Fig. \ref{fig:similarityvscontrastive}). 

\begin{table*}[h]
	\begin{center}
		\resizebox{8cm}{!}{%
			\begin{tabular}{| l | l | l  |}
				\hline
				\textbf{Method} & \textbf{Liver IoU} & \textbf{Shoulder IoU} \\ \hline
				\textbf{Universeg} & 57.75 & 61.3  \\ \hline
				\textbf{PerSAM} & 43.03 & 55.8 \\ \hline
				\textbf{Basic Adapter} & 35 & 51.5 \\ \hline
				\textbf{Classification Adapter} & 57 & 49.1 \\ \hline
				\textbf{Contrastive Adapter} & \textbf{82}  & \textbf{86.6} \\ \hline
			\end{tabular}
		}
	\end{center}
	\caption{Comparison of IoU for segmentation tasks among different methods across liver and shoulder anatomies.}
	\label{table:SegAccuracy}
\end{table*}

\section{Conclusion}
We introduce a method for few-shot multi-label localization and segmentation of medical image volumes, utilizing robust contextual feature vectors extracted from transformer models trained on medical data. These models use a few template image for localization  of all  volumes to produce segmentation or localization outputs. To account for variability within the data and consequently feature vectors which impact heuristic threshold based methods, for initial localization of images, we proposed a contrastive similarity metric learning model, extending it by adopting a multi-label contrastive strategy for enabling tasks with multiple labels. As compared to the state-of-the-art methods for few-shot segmentation/localization not requiring in-domain training, the proposed contrastive adapter approach is superior for both segmentation and localization tasks.

\bibliographystyle{IEEEtran}
\bibliography{samplebibliography}

\end{document}